\documentclass[letterpaper, 10 pt, conference]{ieeeconf}  

\IEEEoverridecommandlockouts                              

\makeatletter
\let\NAT@parse\undefined
\makeatother

\usepackage[
  colorlinks=true,
  linkcolor=cyan,
  urlcolor=black,
  citecolor=green
]{hyperref}
\usepackage{multirow}
\usepackage{multicol}
\usepackage{algpseudocode}
\usepackage{amsmath, amssymb, bm}
\usepackage[ruled,vlined,linesnumbered]{algorithm2e}

\usepackage{graphicx}

\title{\LARGE \bf
SFVO: Decoupled Confidence-Guided Stereo-Flow Visual Odometry with Bidirectional PnP
}

\author{Kai Zhang, Guoyang Zhao, and Jun Ma
\thanks{Kai Zhang and Guoyang Zhao are with the Robotics and Autonomous Systems Thrust, The Hong Kong University of Science and Technology (Guangzhou), Guangzhou 511453, China
(e-mail: kzhang740@connect.hkust-gz.edu.cn; gzhao492@connect.hkust-gz.edu.cn).}
\thanks{Jun Ma is with the Robotics and Autonomous Systems Thrust, The Hong Kong University of Science and Technology (Guangzhou), Guangzhou 511453, China, and also with the Cheng Kar-Shun Robotics Institute, The Hong Kong University of Science and Technology, Hong Kong SAR, China (e-mail: jun.ma@ust.hk). \textit{(Corresponding Author: Jun Ma.)}}
}

\begin{document}

\maketitle
\thispagestyle{empty}
\pagestyle{empty}


\begin{abstract}

Deep learning-based visual odometry (VO) has achieved significant progress, yet most existing methods focus on a monocular approach, which suffers from scale ambiguity. Stereo VO provides real metric by its nature, but remains less studied in deep learning VO due to its high computational cost and modeling complexity. Recent advances in stereo matching and optical flow estimation have made dense visual correspondence increasingly accurate and reliable, but their complementary geometric information has not been fully exploited for VO. In this paper, we present SFVO, a correspondence-driven stereo VO framework that directly builds upon pretrained stereo matching and optical flow models. SFVO exploits pretrained stereo matching and optical flow models to estimate stereo and temporal correspondences. Instead of learning pose directly from images, SFVO maps learned correspondences into geometric constraints and predicts which points are trustworthy.
To improve the reliability of visual correspondence-based geometric constraints, we introduce decoupled confidence maps for rotation and translation. This design better aligns the characteristics of visual correspondence and 6-DoF transformations. Extensive experiments on outdoor and indoor datasets demonstrate that SFVO achieves robust and accurate pose estimation with strong generalization capability.
The code will be released.

\end{abstract}

\section{INTRODUCTION}

Visual odometry (VO) estimates camera motion from consecutive visual observations and is fundamental to robotic navigation \cite{jiang2025semantic}, autonomous driving \cite{zhao2025tsclip}, and aerial robotics. Classical VO typically follows a correspondence-to-geometry paradigm, where visual features are matched across frames and camera motion is recovered from multi-view geometry constraints, which often require intensive engineering efforts. Deep learning has progressively replaced individual components with learned representations and matching models \cite{detone2018superpoint, sarlin2020superglue, teed2020raft_flow}, while also enabling direct end-to-end visual pose estimation. 

Most learning-based VO methods are developed primarily for monocular inputs \cite{zhang2026monoglass3d, ma2024every}, which usually adopt monocular depth estimation as auxiliary task for geometric and image reconstruction supervision \cite{d3vo, deepPatchVSLAM, zhao2025fisheyedepth}. However, monocular vision intrinsically suffers from scale ambiguity. Visual-inertial odometry (VIO) approaches recover metric motion by incorporating inertial measurements \cite{vinsmono, rovio}, but require an additional sensing modality and well-established scale initializations.
Stereo vision resolves scale ambiguity by recovering metric depth from a known baseline. However, learning-based stereo VO remains relatively underexplored, since stereo inputs increase computation and memory cost, while jointly modeling stereo and temporal relationships complicates end-to-end design. Consequently, stereo VO is still commonly based on conventional feature matching and geometric optimization \cite{zhang2024dynpl_stereo, contreras2024dynanav}. Meanwhile, learning-based stereo matching \cite{xu2020aanet_stereo, xu2022attention_stereo} and optical flow \cite{teed2020raft_flow, xu2023unifying_flow} have achieved substantial progress, but their geometric information has not been fully exploited for stereo VO.
This motivates a simpler perspective: stereo matching and optical flow already provide the essential correspondences for frame-to-frame pose estimation. Disparity recovers metric 3D points, while optical flow associates them with 2D observations in adjacent frames, this can be directly formulated as dense 3D--2D constraints. The remaining challenge is to determine which correspondences are reliable and how they should contribute to pose estimation.

\begin{figure}
    \centering
    \includegraphics[width=\linewidth]{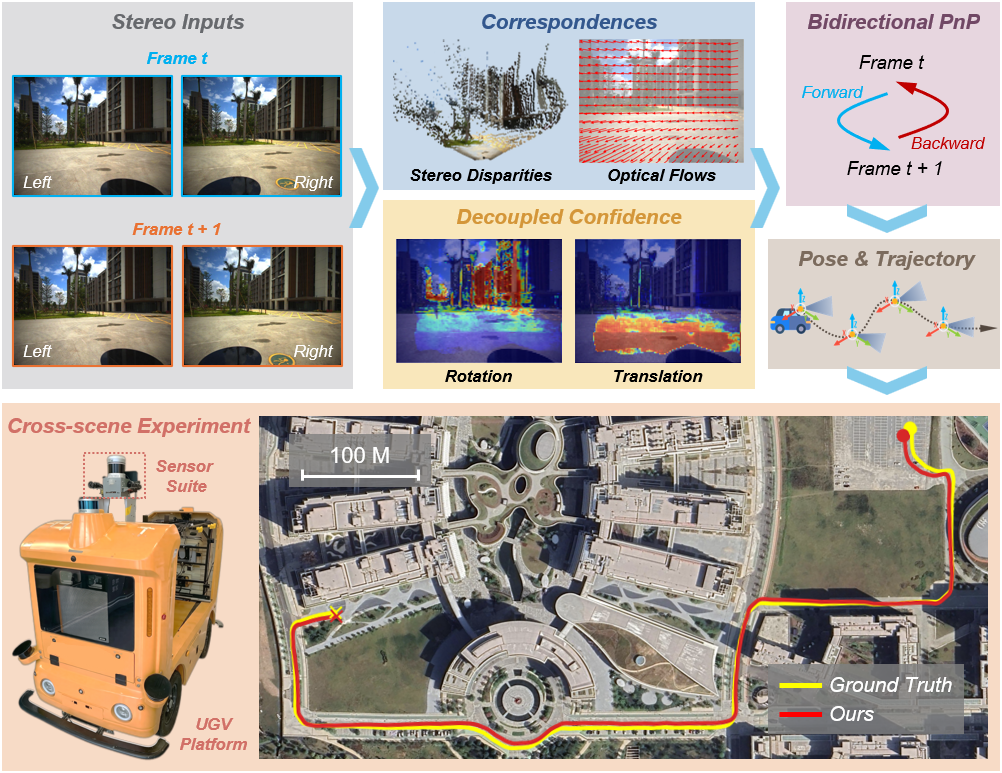}
    \caption{\textbf{Overview of SFVO and its application on an unmanned ground vehicle (UGV).}
    SFVO uses pretrained stereo disparity and optical flow models to establish bidirectional 3D–2D correspondences. Decoupled rotation and translation confidence maps guide a bidirectional PnP solver for frame-to-frame pose estimation. The bottom panel shows the UGV platform and a representative trajectory, demonstrating accurate motion estimation in a real-world environment.}
    \label{fig:intro}
    \vspace{-12pt}
\end{figure}

Based on this perspective, we present SFVO, a correspondence driven deep learning stereo VO framework built upon pretrained stereo matching and optical flow models. As illustrated in Fig.~\ref{fig:intro}, SFVO replaces both heavily engineered correspondence pipelines or black-box pose regression with a modular correspondence-to-geometry formulation. By incorporating pretrained correspondence models, SFVO focuses learning on VO-specific components, namely confidence estimation and geometric pose refinement, thereby reducing the training burden.
We further introduce rotation--translation decoupled confidence maps. Based on the observation that distant points generally provide stronger rotational constraints, while nearby points are more informative for translation, a single confidence value cannot fully represent the geometric utility of each correspondence. SFVO therefore predicts separate rotation and translation confidence channels in both temporal directions and integrates them into a differentiable bidirectional perspective n points (PnP) solver.

The main contributions of this work are summarized as follows:
\begin{itemize}
    \item We propose SFVO, a deep learning-based stereo VO framework built upon pretrained stereo matching and optical flow models, enabling efficient metric pose estimation with reduced training effort.

    \item We formulate stereo VO as a bidirectional correspondence-to-geometry problem, where disparity and optical flow are converted into forward and backward 3D--2D constraints. We integrate Rotation--translation decoupled confidence weighting into a differentiable bidirectional PnP solver.

    \item Experiments on outdoor and indoor benchmarks demonstrate accurate and robust stereo visual odometry performance, with strong generalization to unseen datasets.
\end{itemize}

\section{Related Works}

\subsection{Deep Learning Based Visual Odometry}

Classical VO and visual SLAM systems achieve strong accuracy through carefully engineered modules for feature extraction, data association, outlier rejection, and geometric optimization \cite{forster2014svo, vinsmono, campos2021orbslam3}. Learning-based VO reduces such manual design by learning visual representations or motion estimation directly from data.
DeepVO \cite{wang2017deepvo} directly regresses camera motion from image sequences, while later methods such as DPV-SLAM \cite{deepPatchVSLAM}, DF-VO \cite{zhan2020visual_dfvo}, and D3VO \cite{d3vo} incorporate monocular depth and image reconstruction for geometric supervision. Visual-inertial methods, including UnVIO \cite{unvio_ijcai2021} and BotVIO \cite{wei2025botvio}, incorporate learned visual and inertial features. However, these approaches, including learned VIO methods, remain primarily based on monocular vision and therefore depend on estimated depth or additional scale recovery.

Stereo vision directly provides metric depth from a known baseline, but learning-based stereo VO remains less explored due to increased modeling complexity and computation cost. SFVO performs frame-to-frame pose estimation by converting pretrained stereo and optical-flow correspondences directly into geometric constraints.

\subsection{Deep Learning Visual Correspondence Estimation}

Recent deep learning correspondence models provide dense alternatives to handcrafted feature matching. For stereo matching, AANet \cite{xu2020aanet_stereo} improves efficiency with lightweight cost aggregation, ACVNet \cite{xu2022attention_stereo} introduces attention-based cost volumes, while RAFT-Stereo \cite{lipson2021raft_stereo} and CREStereo \cite{li2022practical_stereo} employ recurrent refinement for accurate and robust disparity estimation. These methods provide metric depth for calibrated stereo systems.
For optical flow, RAFT \cite{teed2020raft_flow}, FlowFormer \cite{huang2022flowformer_flow}, CRAFT \cite{sui2022craft_flow}, and UniMatch \cite{xu2023unifying_flow} achieve strong optical flow estimation through recurrent refinement or attention-based matching. Combined with stereo-derived depth, optical flow naturally establishes dense 3D--2D constraints for frame-to-frame pose estimation.

Despite substantial progress in stereo matching and optical flow, these models are primarily optimized for correspondence accuracy rather than pose estimation. Their dense outputs may still contain unreliable matches caused by occlusions, dynamic objects, or estimation errors. SFVO therefore focuses on converting correspondences into pose constraints and learning how each correspondence should contribute to pose optimization.

\section{Methodology}

\begin{figure*}[h!]
    \centering
    \includegraphics[width=1\linewidth]{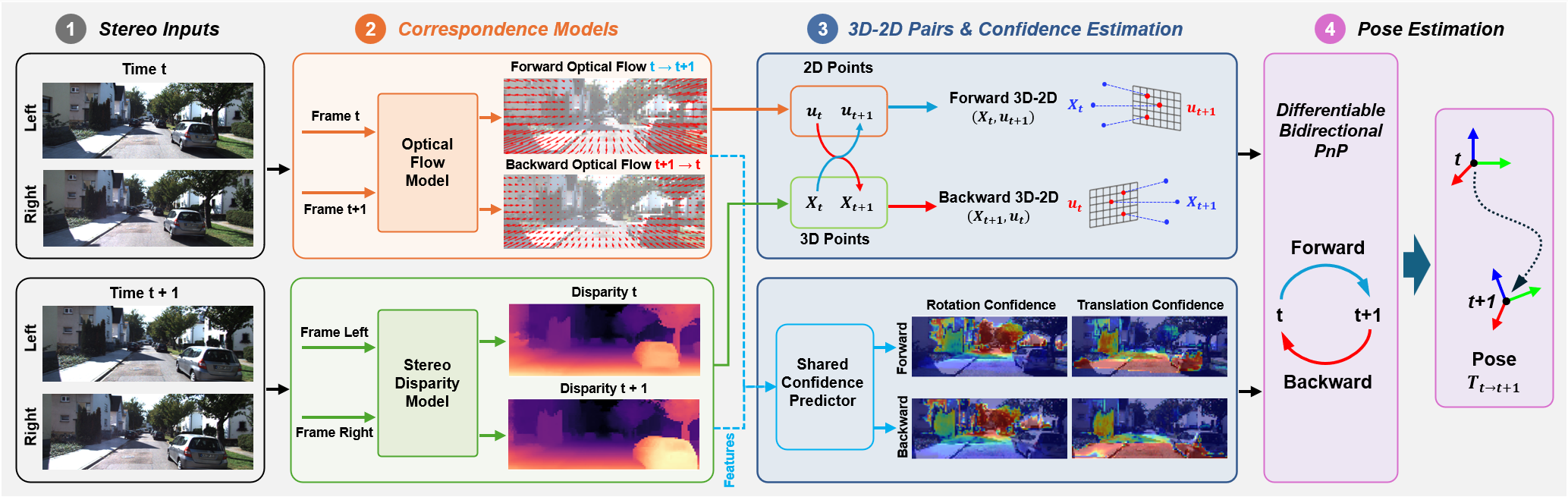}
    \caption{\textbf{SFVO structure.} The input consists of two stereo image frames, stereo disparities are estimated by stereo matching model, forward and backward optical flow are estimated by optical flow model. The confidence estimator predicts translation-rotation-decoupled confidence maps from features extracted by optical flow and disparity models, which serve as weighting factors in the differentiable PnP pose solver.}
    \label{fig:frame}
    \vspace{-16pt}
\end{figure*}

\subsection{System Overview}

The proposed SFVO framework is illustrated in Fig.~\ref{fig:frame}. Given two consecutive rectified stereo frames at timestamps $t$ and $t+1$, SFVO estimates the relative camera transformation $\mathbf{T}_{t\rightarrow t+1}\in SE(3)$.

Each stereo pair is processed by a pretrained stereo matching network to estimate left to right disparities $\delta_t$ and $\delta_{t+1}$. With focal length $f_x$ and stereo baseline $b$, disparity is converted into metric depth as
\begin{equation}
    Z_k(\mathbf{u})
    =
    \frac{f_x b}{\delta_k(\mathbf{u})},
    \qquad k \in \{t,t+1\}
    \label{eq:disp_to_depth}
\end{equation}

Meanwhile, the two left images are processed by a pretrained optical flow network to estimate forward and backward temporal correspondences, $t\rightarrow t+1$ and $t+1\rightarrow t$. Combined with the stereo-derived 3D points, these flow fields form bidirectional 3D--2D correspondences.

Features from the stereo and optical flow networks are passed to a confidence estimation module, which predicts separate rotation and translation confidence weights for both temporal directions.

Finally, the bidirectional 3D--2D correspondences and their confidence weights are jointly processed by a weighted bidirectional PnP solver to estimate the relative camera pose.

\subsection{Correspondence Estimation Model}

For both stereo disparity and optical flow estimation, we adopt the pretrained UniMatch model \cite{xu2023unifying_flow}. Using a unified correspondence backbone simplifies the architecture and allows SFVO to directly reuse pretrained dense matching representations.

UniMatch establishes correspondences through global feature matching rather than task-specific regression decoders \cite{teed2020raft_flow, huang2022flowformer_flow}. It computes pairwise feature similarities, converts them into matching distributions with softmax, and obtains correspondence displacements from the estimated target pixel coordinates. This coordinate-based formulation supports different image resolutions and aspect ratios while avoiding heavy task-specific decoding.

In SFVO, we retain only the correspondence output at $1/8$ of the original image resolution and remove subsequent upsampling stages. Since our objective is pose estimation rather than high-resolution disparity or flow reconstruction, maintaining low resolution provides sufficiently dense geometric constraints while substantially reducing GPU memory consumption during end-to-end training.

\subsection{Decoupled Confidence Prediction}
\label{sec:method_conf}

Dense correspondences can be corrupted by occlusions, dynamic objects, textureless regions, and matching ambiguities, making confidence weighting essential for robust pose estimation. Most existing methods assign a single confidence value to each correspondence, implicitly assuming equal reliability for rotation and translation. However, their geometric contributions are inherently different.

Consider a 3D point $\mathbf{X}=[X,Y,Z]^T$ projected to normalized image coordinates $\mathbf{u}=[u_x,u_y]^T=[X/Z,Y/Z]^T$. Under small translational velocity $\mathbf{v}$ and angular velocity $\boldsymbol{\omega}$, the induced image motion is
\begin{equation}
    \dot{\mathbf{u}}
    =
    \mathbf{J}_{\mathrm{trans}}(\mathbf{u},Z)\mathbf{v}
    +
    \mathbf{J}_{\mathrm{rot}}(\mathbf{u})\boldsymbol{\omega}
    \label{eq:motion_field}
\end{equation}
where
\begin{equation}
    \mathbf{J}_{\mathrm{trans}}
    =
    \frac{1}{Z}
    \begin{bmatrix}
        -1 & 0 & u_x \\
        0 & -1 & u_y
    \end{bmatrix}
\end{equation}
and
\begin{equation}
    \mathbf{J}_{\mathrm{rot}}
    =
    \begin{bmatrix}
        u_x u_y & -(1+u_x^2) & u_y \\
        1+u_y^2 & -u_x u_y & -u_x
    \end{bmatrix}
\end{equation}
The translational component decreases with depth $Z$, whereas the rotational component is depth-independent. Thus, nearby points generally provide stronger translational constraints, while distant points remain the same informativeness for rotation.

This distinction is further amplified by stereo depth uncertainty. Given disparity $\delta$, a small disparity error $\Delta\delta$ induces depth error $\Delta Z$ as
\begin{equation}
    \Delta Z
    \approx
    \frac{Z^2}{f_x b}\Delta\delta
    \label{eq:depth_uncertainty}
\end{equation}
which shows that absolute depth error grows quadratically with distance. Consequently, distant points may provide uncertain metric constraints for translation while retaining useful rotational information, which cannot be represented effectively by a shared confidence value.

SFVO therefore predicts separate confidence maps for rotation and translation. Given fused stereo and optical-flow features $\mathbf{F}_d$, the confidence estimator outputs
\begin{equation}
    \left[ \mathbf{W}^{r}, \mathbf{W}^{t} \right]
    =
    \mathrm{Sigmoid}\left(g(\mathbf{F}_{d})\right)
    \label{eq:confidence_prediction}
\end{equation}
where $g$ is the confidence estimator layers consists of 3 transformer encoder and decoder layers, and $\mathrm{Sigmoid}$ constrains the confidence values to $(0,1)$. Applying the estimator in both forward and backward temporal directions yields four maps:
$\mathbf{W}^{r}_{f}$, $\mathbf{W}^{t}_{f}$,
$\mathbf{W}^{r}_{b}$, and $\mathbf{W}^{t}_{b}$.

This decoupling allows SFVO to retain distant correspondences for rotation while assigning stronger translation confidence to nearby points with informative parallax.

\subsection{Differentiable Bidirectional PnP}

To integrate decoupled confidence weights, we decompose the PnP problem into two differentiable subproblems with closed-form updates. Translation and rotation are estimated alternately; at each iteration, the rotation is updated using the current translation estimate, after which the translation is updated using the newly estimated rotation.

\subsubsection{Rotation Estimation}

For rotation estimation, we adopt Wahba's problem \cite{wahba}, whose objective is to find the rotation that best aligns two sets of corresponding unit vectors:
\begin{equation}
    L=\sum_{i=1}^N w_i\|\mathbf{c}_i-\mathbf{R}\mathbf{a}_i\|^2
\end{equation}
where $L$ denotes the weighted sum of squared alignment errors, $w_i$ is the confidence weight of correspondence $i$, $\mathbf{a}_i$ and $\mathbf{c}_i$ are corresponding unit vectors expressed in two reference frames. In our problem setup, $\mathbf{a}_i$ is in frame $t$, $\mathbf{c}_i$ is in frame $t+1$, $\mathbf{R}$ represents the rotation from frame $t$ to frame $t+1$.

For the forward direction, the inputs consist of a 3D point
$\mathbf{X}_t\in\mathbb{R}^{3\times1}$ in frame $t$, its corresponding normalized image coordinate
$\mathbf{u}_{t+1}=[u_x,u_y,1]^T$ in frame $t+1$, and the forward rotation confidence
$w_f^r\in(0,1)$.

To formulate the forward Wahba constraint, we first define the source direction as
$\mathbf{a}_f=\mathrm{norm}_2(\mathbf{X}_t)$, where $\mathrm{norm}_2(\cdot)$ denotes $\ell_2$ normalization. The normalized image coordinate is converted into a unit bearing vector $\mathbf{b}_{t+1} = \frac{\mathbf{u}_{t+1}}{\|\mathbf{u}_{t+1}\|}$.
The bearing vector represents the viewing direction along the transformed 3D point. Using the current pose estimate, the point is transformed from frame $t$ to frame $t+1$ as
\begin{equation}
    \hat{\mathbf{X}}_{t+1} = \mathbf{R}\mathbf{X}_t + \mathbf{t}
\end{equation}
Its orthogonal projection onto the unit bearing direction
$\mathbf{b}_{t+1}$ is
$(\mathbf{b}_{t+1}^T\hat{\mathbf{X}}_{t+1})\mathbf{b}_{t+1}$.
After removing the current translation estimate, the target direction associated with the rotational component is defined as
\begin{equation}
    \mathbf{c}_f =
    \mathrm{norm}_2 \left(
    (\mathbf{b}_{t+1}^T \hat{\mathbf{X}}_{t+1}) \mathbf{b}_{t+1} - \mathbf{t}
    \right)
\end{equation}
Thus, $\mathbf{c}_f\in\mathbb{R}^{3\times1}$ represents the estimated direction toward which $\mathbf{R}\mathbf{X}_t$ should align. Together with
$\mathbf{a}_f$ and the rotation weight $w_f^r$, this defines the forward Wahba constraint.

For the backward direction, the inputs are the 3D point
$\mathbf{X}_{t+1}$, its corresponding normalized image coordinate
$\mathbf{u}_t$, and the backward rotation confidence $w_b^r$. Since the metric 3D point is already available in frame $t+1$, no explicit projection-depth estimation is required. The corresponding source and target directions are
\begin{equation}
    \mathbf{a}_b = \mathbf{b}_b,\quad
    \mathbf{c}_b=\mathrm{norm}_2(\mathbf{X}_{t+1} - \mathbf{t})
\end{equation}

To jointly estimate the rotation from both forward and backward constraints, we construct the weighted cross covariance matrix
\begin{equation}
\mathbf{H} = \sum_i w^{r}_{f,i}\mathbf{c}_{f,i}\mathbf{a}_{f,i}^{T} + 
\sum_i w^{r}_{b,i}\mathbf{c}_{b,i}\mathbf{b}_{b,i}^{T}
\end{equation}
We then perform singular value decomposition,
$\mathbf{H}=\mathbf{U}\mathbf{\Sigma}\mathbf{V}^T$.
The weighted least-squares solution to Wahba's problem is
\begin{equation}
\mathbf{R}
=
\mathbf{U}
\operatorname{diag}(1,1,\det(\mathbf{U}\mathbf{V}^{T}))
\mathbf{V}^{T}
\end{equation}
where the determinant correction guarantees that
$\mathbf{R}\in SO(3)$ and prevents the solution from becoming a reflection.

\subsubsection{Translation Estimation}

Given the updated rotation, translation is estimated by minimizing the weighted bearing-perpendicular residuals. For a unit bearing vector $\mathbf{b}$, the projection matrix
\begin{equation}
    \mathbf{P} = \mathbf{I}_{3\times3} - \mathbf{b}\mathbf{b}^T
\end{equation}
projects a vector onto the plane perpendicular to $\mathbf{b}$. If a transformed 3D point $\mathbf{v}$ is perfectly aligned with its observed position, its perpendicular component satisfies $\mathbf{P}\mathbf{v}=0$.

For the forward direction, the weighted translation objective is therefore defined as
\begin{equation}
    \min_{\mathbf{t}}
    {\sum_{i}w_i^t
    \|\mathbf{P}_{f,i}(\mathbf{R}\mathbf{X}_{t,i}
    + \mathbf{t})\|^2}
\end{equation}
with a geometric constraint for each correspondence
\begin{equation}
    \mathbf{P}_{f,i}
    (\mathbf{R}\mathbf{X}_{t,i} + \mathbf{t})=0
\end{equation}
which can be rearranged as
\begin{equation}
    \mathbf{P}_{f,i}\mathbf{t}
    =
    -\mathbf{P}_{f,i}\mathbf{R}\mathbf{X}_{t,i}
\end{equation}

Accumulating all weighted constraints yields the forward linear system
$\mathbf{A}_f\mathbf{t}=\mathbf{r}_f$, where
\begin{equation}
\begin{split}
    \mathbf{A}_f
    &= \sum_i w^{t}_{f,i}\mathbf{P}_{f,i} \\
    \mathbf{r}_f
    &= -\sum_i w^{t}_{f,i}
    \mathbf{P}_{f,i}\mathbf{R}\mathbf{X}_{t,i}
\end{split}
\end{equation}

For the backward direction, the inverse-transformed 3D point should align with its bearing vector in frame $t$. The corresponding geometric constraint is
\begin{equation}
    \mathbf{P}_{b,i}\mathbf{R}^T
    (\mathbf{X}_{t+1,i} - \mathbf{t})=0
\end{equation}
Expressing the resulting normal equation in the coordinate frame $t+1$ gives
\begin{equation}
\begin{split}
    \mathbf{A}_b
    &=
    \mathbf{R}
    \left(
    \sum_i w^{t}_{b,i}\mathbf{P}_{b,i}
    \right)
    \mathbf{R}^T \\
    \mathbf{r}_b
    &=
    \mathbf{R}
    \left(
    \sum_i w^{t}_{b,i}
    \mathbf{P}_{b,i}\mathbf{R}^T\mathbf{X}_{t+1,i}
    \right)
\end{split}
\end{equation}

Combining the forward and backward linear systems, the translation is obtained as
\begin{equation}
    \mathbf{t}
    =
    (\mathbf{A}_f+\mathbf{A}_b+\epsilon\mathbf{I})^{-1}
    (\mathbf{r}_f+\mathbf{r}_b)
\end{equation}
where $\epsilon\mathbf{I}$ is a small regularization term that improves numerical stability.

\begin{algorithm}[h!]
\caption{Bidirectional Decoupled Weighted PnP Solver}
\label{alg:bidir_pnp}
\KwIn{
Forward 3D--2D correspondences $(\mathbf{X}_t,\mathbf{u}_{t+1})$, 
backward 3D--2D correspondences $(\mathbf{X}_{t+1},\mathbf{u}_t)$, 
rotation weights $(\mathbf{w}^{r}_f,\mathbf{w}^{r}_b)$, 
translation weights $(\mathbf{w}^{t}_f,\mathbf{w}^{t}_b)$, 
iterations $K$
}
\KwOut{Pose $(\mathbf{R},\mathbf{t})$ from frame $t$ to frame $t+1$}

Normalize 2D points to unit bearing vectors
$\mathbf{b} = \frac{\mathbf{u}}{\|\mathbf{u}\|}$

Initialize $\mathbf{R}\leftarrow \mathbf{I}$,
$\mathbf{t}\leftarrow \mathbf{0}$\;

\For{$k=1$ \KwTo $K$}{

\textbf{Rotation update}\;

Forward rotation constraints $\mathbf{a}_f,\, \mathbf{c}_f$

Backward rotation constraints $\mathbf{a}_b,\, \mathbf{c}_b$

Construct weighted cross-covariance matrix $\mathbf{H}$

Solve $\mathbf{H}=\mathbf{U}\mathbf{\Sigma}\mathbf{V}^{T}$
and update $\mathbf{R}$

\textbf{Translation update}\;

Forward translation constraints $\mathbf{A}_f,\,\mathbf{r}_f$

Backward translation constraints $\mathbf{A}_b,\,\mathbf{r}_b$

Solve $\mathbf{A}\mathbf{t}=\mathbf{r}$
and update $\mathbf{t}$

}
\Return{$\mathbf{R},\mathbf{t}$}
\end{algorithm}

Algorithm~\ref{alg:bidir_pnp} summarizes the alternating rotation and translation updates. During training, rotation and translation are initialized with $\mathbf{R}=\mathbf{I}$ and $\mathbf{t}=\mathbf{0}$. At test time, it can utilize the relative motion estimated for the previous frame pair as a warm start.

The solver is implemented with standard batched linear algebra operations. These operations can be programmed efficiently with PyTorch and support gradient backpropagation, allowing pose supervision to jointly optimize the correspondence and confidence prediction modules while retaining efficient computation.

\subsection{Loss Functions}

The optical flow and stereo networks are initialized with pretrained UniMatch weights. Since VO datasets generally lack ground-truth flow and disparity, correspondence estimation is supervised by image reconstruction. The correspondence loss $\mathcal{L}_{\mathrm{corr}}$ is computed as the SSIM loss between the reference image and the target image warped using the estimated correspondence.

Pose estimation is supervised by the ground-truth transformation. Translation is optimized using an $\ell_1$ loss:
\begin{equation}
    \mathcal{L}_{\mathrm{trans}}
    =
    \|\mathbf{t}_{\mathrm{pred}}-\mathbf{t}_{\mathrm{gt}}\|_1
\end{equation}

For rotation, we define
$\mathbf{R}_{\mathrm{err}}=\mathbf{R}_{\mathrm{pred}}^T\mathbf{R}_{\mathrm{gt}}$
and minimize its angular error:
\begin{equation}
    \theta =
    \cos^{-1}\left(
    \frac{\mathrm{tr}(\mathbf{R}_{\mathrm{err}})-1}{2}
    \right),
    \quad
    \mathcal{L}_{\mathrm{rot}}=\mathrm{mean}(\theta)
\end{equation}
For very small angles, the norm of the skew-symmetric component of $\mathbf{R}_{\mathrm{err}}$ is used to approximate $\theta$ for numerical stability.

To prevent the confidence maps from collapsing toward uniformly small values, we introduce a regularization loss with reference $r=0.15$
\begin{equation}
    \mathcal{L}_{\mathrm{conf}}^{\mathrm{ref}}
    =
    \frac{1}{N}\sum_{i=1}^{N}
    \mathrm{ReLU}(r-w_i),
    \quad r=0.15
\end{equation}
This one-sided regularization with $\mathrm{ReLU}$ penalizes confidence values below $r$, while leaving larger values unconstrained.

For translation confidence, each valid correspondence produces an implicit translation estimate $\mathbf{t}_i$ using the ground-truth rotation and depth. Their confidence-weighted estimate is compared with the ground truth:
\begin{equation}
    \mathcal{L}_{\mathrm{conf}}^{t} = \log\left( 1 + \sum_i \bar{w}_{i}^{t} \left\| \mathbf{t}_i -
    \mathbf{t}_{\mathrm{gt}} \right\| \right)
\end{equation}
where $\bar{w}_i^t=w_i^t/\sum_jw_j^t$ is the normalized translation confidence, natural log is applied to regulate large errors. For rotation confidence, we use the same bearing-vector formulation as the Wahba solver:
\begin{equation}
    \mathcal{L}_{\mathrm{conf}}^{r}
    =
    \sum_i \bar{w}_i^r
    \cos^{-1}
    \left(
    (\mathbf{R}_{\mathrm{gt}}\mathbf{a}_i)^T\mathbf{c}_i
    \right)
\end{equation}
These confidence losses are applied in both forward and backward directions, with
\begin{equation}
    \mathcal{L}_{\mathrm{conf}}
    =
    \mathcal{L}_{\mathrm{conf}}^{\mathrm{ref}}
    +
    \mathcal{L}_{\mathrm{conf}}^{t}
    +
    \mathcal{L}_{\mathrm{conf}}^{r}
\end{equation}

The overall objective equally weights correspondence, pose, and confidence supervision:
\begin{equation}
    \mathcal{L}
    =
    \mathcal{L}_{\mathrm{corr}}
    +
    \mathcal{L}_{\mathrm{trans}}
    +
    \mathcal{L}_{\mathrm{rot}}
    +
    \mathcal{L}_{\mathrm{conf}}
\end{equation}

\section{Experiments}

\subsection{Experimental Setup}

SFVO is implemented with PyTorch and initialized with publicly released UniMatch weights pretrained on mixed optical flow and stereo disparity datasets. All models are trained end-to-end on 4 NVIDIA GeForce RTX 4090 GPUs with a learning rate of $1\times10^{-4}$. We evaluate SFVO on KITTI Odometry \cite{kitti} and EuRoC MAV \cite{euroc}, covering both outdoor driving and indoor aerial scenarios.

For EuRoC MAV, we use MH02 (E), MH04 (D), and V103 (D) for evaluation and the remaining sequences for training, with 5 training epochs.

For KITTI, following existing VO works \cite{unvio_ijcai2021, wei2025botvio}, sequences 00--08 are used for training and 09--10 for evaluation. The model is trained on KITTI for 5 epochs.

To assess cross-dataset generalization, we evaluate our trained model on an sequence collected from a UGV, the platform and trajectory is introduced in Fusion Portable V2 dataset \cite{wei2025fusionportablev2}, this dataset is not used in any training procedure.

\subsection{Experimental Results}

\begin{table*}[h!]
    \centering
    \caption{Quantitative comparison on the EuRoC MAV dataset.}
    \label{tab:euroc}
    \renewcommand{\arraystretch}{1.2}
    \begin{tabular}{l c c c c c c c c c c}
        \hline
        \multirow{2}{*}{Method} & \multirow{2}{*}{Sensor}
        & \multicolumn{3}{c}{MH02 (E)}
        & \multicolumn{3}{c}{MH04 (D)}
        & \multicolumn{3}{c}{V103 (D)} \\
        \cline{3-5} \cline{6-8} \cline{9-11}
        & & ATE (m) & RPE$_t$ (m) & RPE$_r$ (deg)
        & ATE (m) & RPE$_t$ (m) & RPE$_r$ (deg)
        & ATE (m) & RPE$_t$ (m) & RPE$_r$ (deg) \\
        \hline
        DPVO & Mono
        & 0.105 & \underline{0.007} & 0.586
        & 0.177 & \underline{0.007} & \underline{0.685}
        & \underline{0.126} & \underline{0.017} & \underline{1.312} \\
        BotVIO & Mono+IMU
        & 0.190 & - & -
        & \underline{0.150} & - & -
        & 0.200 & - & - \\
        ORB-SLAM3 & Stereo
        & \textbf{0.037} & 0.024 & \underline{0.548}
        & \textbf{0.112} & 0.054 & 0.701
        & 0.199 & 0.040 & 2.326 \\
        SFVO (Ours) & Stereo
        & \underline{0.087} & \textbf{0.002} & \textbf{0.033}
        & 0.217 & \textbf{0.006} & \textbf{0.044}
        & \textbf{0.112} & \textbf{0.009} & \textbf{0.160} \\
        \hline
    \end{tabular}
\end{table*}

We compare SFVO with three representative VO methods. DPVO is a learning-based monocular VO method with bundle-adjustment-based pose refinement, while BotVIO is a monocular visual--inertial method that estimates frame-to-frame motion using visual and inertial measurements without backend optimization. ORB-SLAM3 is a conventional feature-based SLAM framework, for a fair comparison, we evaluate with stereo mode and loop closure disabled.

We use absolute trajectory error (ATE) and relative pose error (RPE) for evaluation. ATE measures global trajectory discrepancy after alignment, while RPE measures frame-to-frame translation and rotation errors, denoted as $\mathrm{RPE}_t$ and $\mathrm{RPE}_r$, respectively. Since DPVO and BotVIO are monocular methods, scale alignment is applied to their trajectories, SFVO and stereo ORB-SLAM3 use rigid alignment only.

\begin{figure}[h!]
    \centering
    \includegraphics[width=\linewidth]{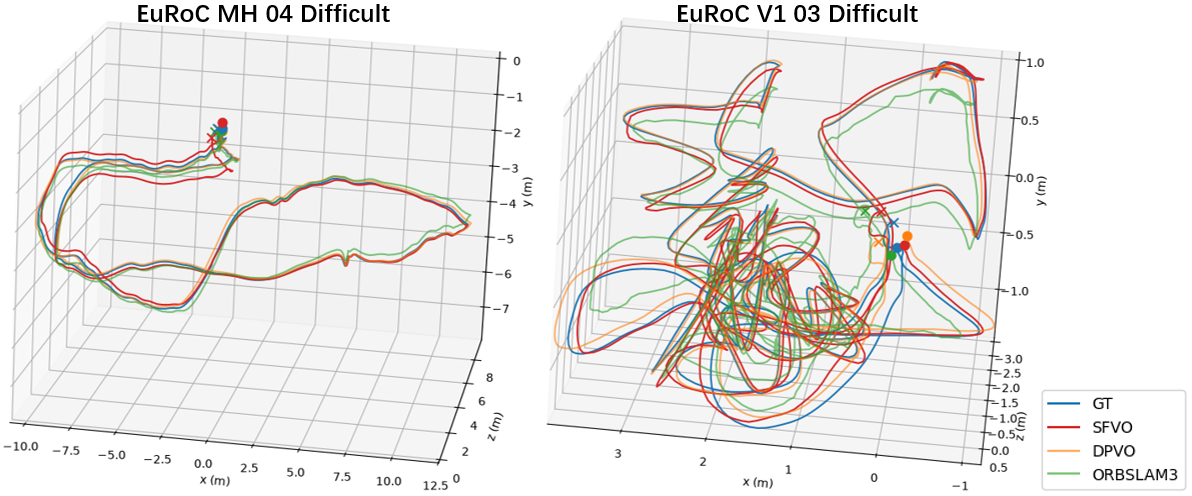}
    \caption{\textbf{Trajectory comparisons on the EuRoC MAV dataset.}
    Results are shown for sequences MH04 (D) (left) and V103 (D) (right).}
    \label{fig:euroc_plot}
\end{figure}

Qualitative results on EuRoC are shown in Fig.~\ref{fig:euroc_plot}, metric comparisons are shown in Table~\ref{tab:euroc}. BotVIO trajectory is not plotted because pretrained EuRoC weights are unavailable. SFVO achieves the best result in seven of nine metrics and consistently obtains the lowest translational and rotational RPE on all three sequences, demonstrating accurate frame-to-frame motion estimation. On V103 (D), SFVO also achieves the lowest ATE. The relatively higher ATE on MH04 (D), despite low RPE, indicates accumulated drift in the absence of backend trajectory optimization.

\begin{figure}[h]
    \centering
    \includegraphics[width=\linewidth]{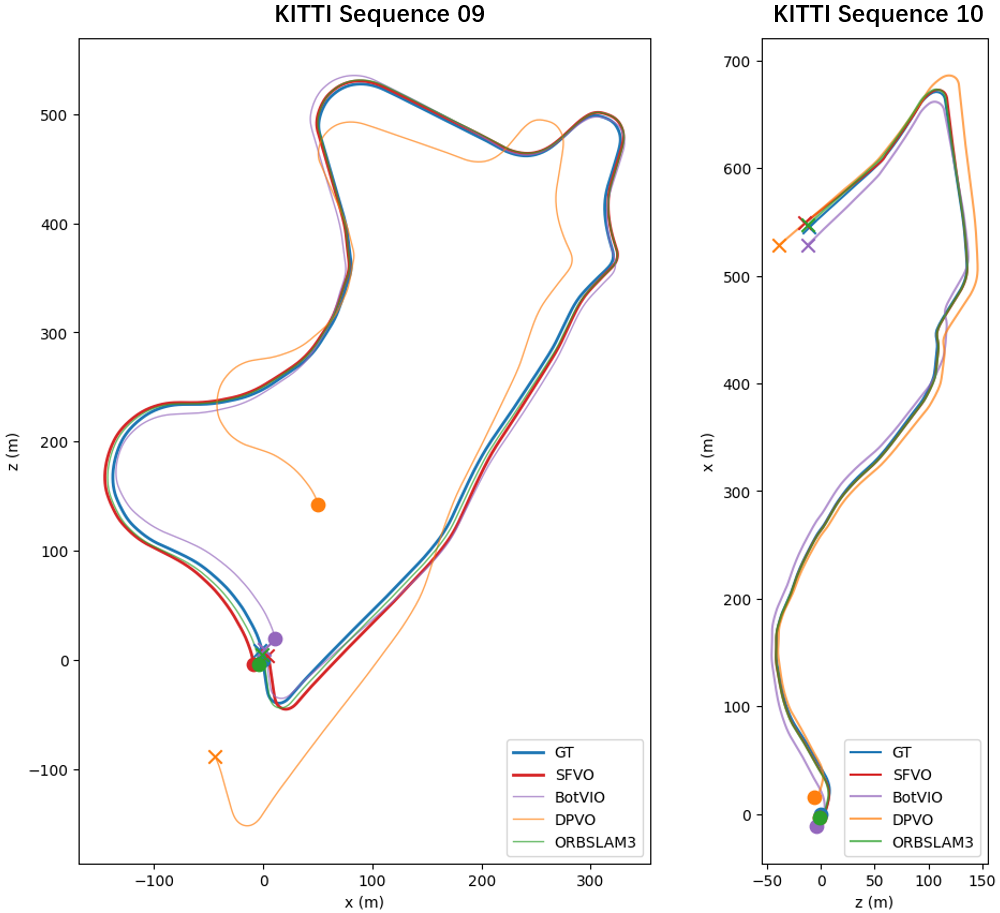}
    \caption{\textbf{Trajectory comparisons on the KITTI odometry dataset.}
    Results are shown for sequences 09 (left) and 10 (right).}
    \label{fig:kitti_plot}
\end{figure}

\begin{table}[h]
    \centering
    \caption{Quantitative comparison on KITTI sequences 09 and 10.}
    \label{tab:kitti}
    \renewcommand{\arraystretch}{1.2}
    \begin{tabular}{c c c c c}
        \hline
        Method & Sequence & ATE (m) & RPE$_t$ (m) & RPE$_r$ (deg) \\
        \hline
        DPVO & \multirow{4}{*}{09} & 74.868 & 0.531 & \textbf{0.037} \\
        BotVIO & & 9.942 & 0.082 & 0.407 \\
        ORB-SLAM3 & & \textbf{2.077} & \underline{0.017} & 0.044 \\
        SFVO (Ours) & & \underline{9.830} & \textbf{0.016} & \underline{0.039} \\
        \hline
        DPVO & \multirow{4}{*}{10} & 13.602 & 0.166 & \textbf{0.044} \\
        BotVIO & & 13.417 & 0.084 & 0.364 \\
        ORB-SLAM3 & & \textbf{1.408} & \underline{0.019} & 0.050 \\
        SFVO (Ours) & & \underline{2.835} & \textbf{0.017} & \underline{0.046} \\
        \hline
    \end{tabular}
\end{table}

Qualitative results on KITTI are presented in Fig.~\ref{fig:kitti_plot}, metric comparisons are shown in Table~\ref{tab:kitti}. SFVO achieves the lowest translational RPE on both sequences and the second-best rotational RPE and ATE, with errors close to the best result. It also substantially outperforms the monocular DPVO and BotVIO baselines in ATE. Despite performing only frame-to-frame estimation, SFVO remains competitive with stereo ORB-SLAM3, which benefits from keyframe management and local window optimization.

\begin{figure}[h]
    \centering
    \includegraphics[width=\linewidth]{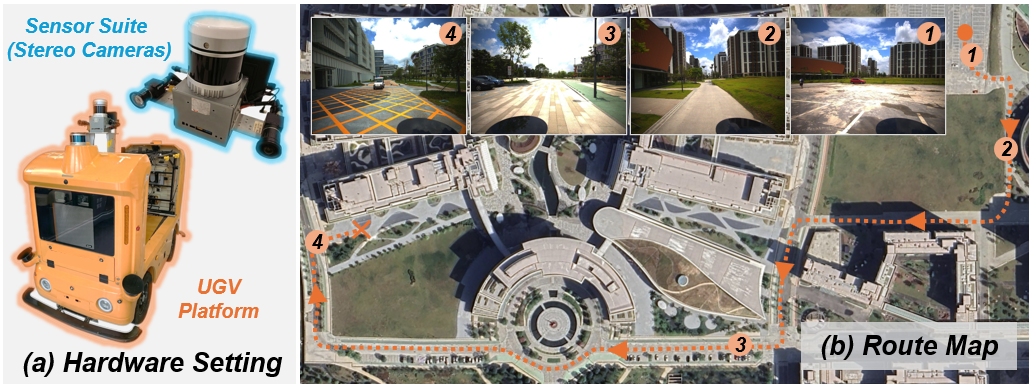}
    \caption{\textbf{UGV platform and sample images from the campus sequence.}
    The data-collection platform is shown on the left, and representative image samples are shown on the right.}
    \label{fig:ugv_platform}
    \vspace{-12pt}
\end{figure}

We further evaluate cross-dataset generalization on an unseen UGV sequence collected with industrial stereo RGB cameras. Ground truth is provided by RTK-enabled GNSS. The sequence is recorded inside a campus region, starts from parking lot and ends in front of a laboratory. The scene appearance differs from KITTI, while the vehicle motion speed is also slower. The UGV and some image samples are shown in Fig.~\ref{fig:ugv_platform}.

\begin{figure}[h!]
    \centering
    \includegraphics[width=\linewidth]{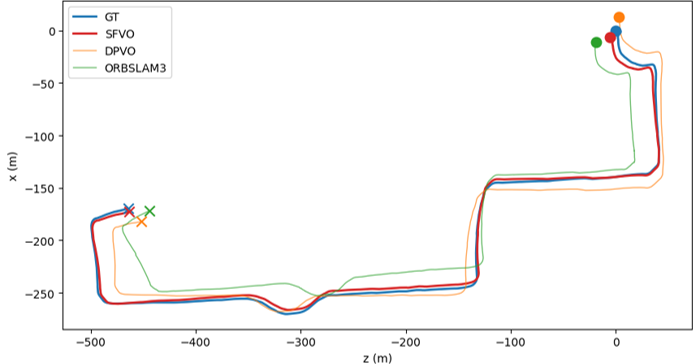}
    \caption{\textbf{Trajectory comparison on the unseen UGV campus sequence.}
    This sequence is used to evaluate generalization to a new platform, motion pattern, and environment.}
    \label{fig:fusion_plot}
\end{figure}

\begin{table}[h!]
    \centering
    \caption{Quantitative comparison on the unseen UGV campus sequence.}
    \label{tab:fusion}
    \renewcommand{\arraystretch}{1.2}
    \begin{tabular}{c c c c c}
        \hline
        Method & Sensor & ATE (m) & RPE$_t$ (m) & RPE$_r$ (deg) \\
        \hline
        DPVO & Mono & 12.798 & 0.052 & \textbf{0.247} \\
        ORB-SLAM3 & Stereo & 12.282 & 0.090 & 0.784 \\
        SFVO (Ours) & Stereo & \textbf{4.906} & \textbf{0.043} & 0.312 \\
        \hline
    \end{tabular}
\end{table}

As shown in Fig.~\ref{fig:fusion_plot} and Table~\ref{tab:fusion}, SFVO achieves the lowest ATE and translational RPE on this unseen sequence, reducing ATE by approximately 60\% relative to the baseline. This demonstrates strong generalization across camera platforms, motion patterns, and environments. Although DPVO obtains a slightly lower rotational RPE, SFVO provides substantially better global trajectory accuracy.

\subsection{Ablation Studies}

\subsubsection{Decoupled Confidence}

\begin{figure}[h]
    \centering
    \includegraphics[width=\linewidth]{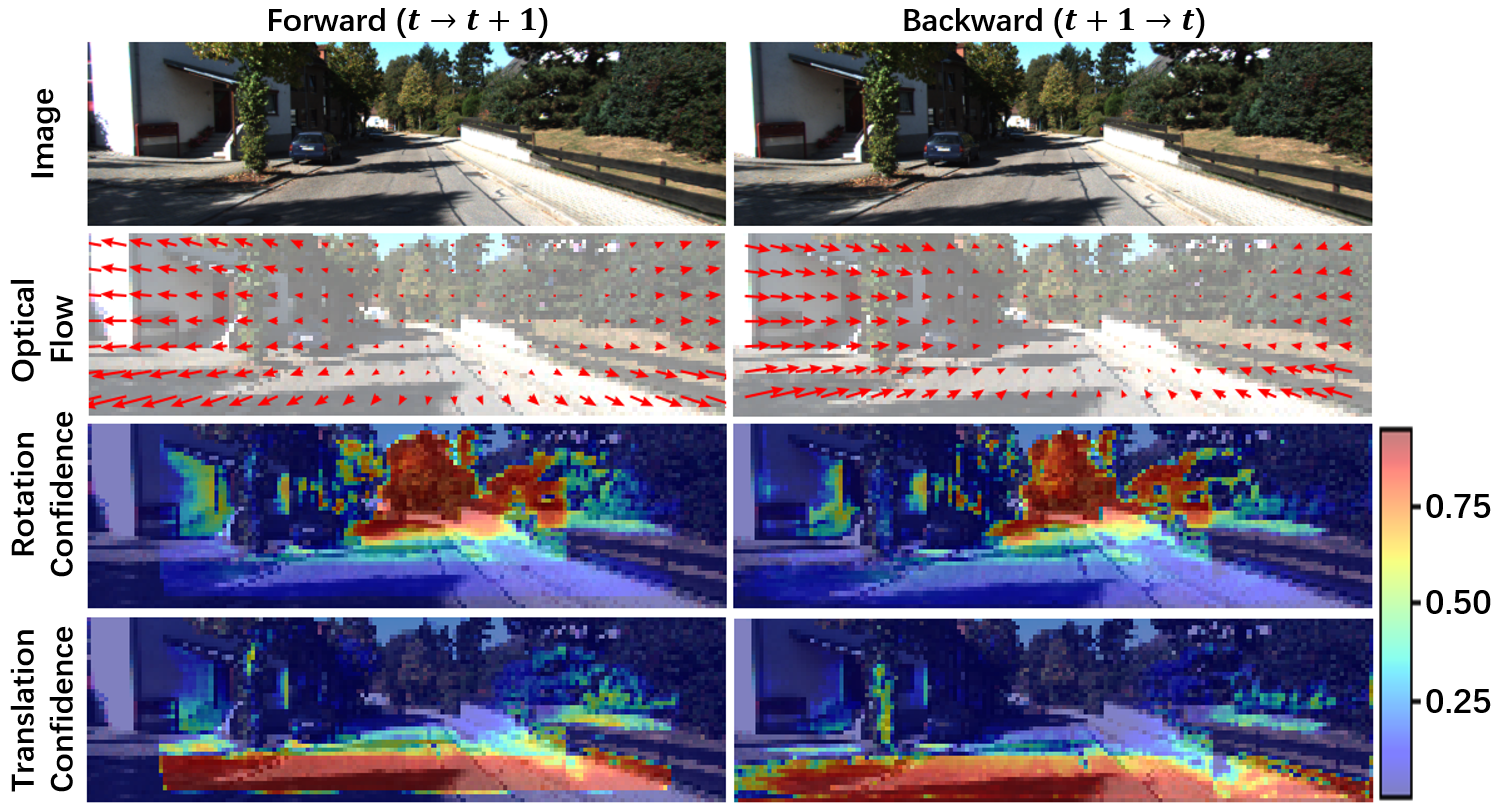}
    \caption{\textbf{Visualization of rotation--translation decoupled confidence maps.}
    Distant regions tend to receive higher rotation confidence, while nearby regions are assigned higher translation confidence, consistent with their different geometric contributions to pose estimation.}
    \label{fig:conf_flow}
\end{figure}

The rotation--translation decoupled confidence design is a key component of SFVO. As shown in Fig.~\ref{fig:conf_flow}, the appearance of translation and rotation confidence maps presents a clear depth-dependent pattern, while the model is trained without explicit depth-based supervision. Distant regions generally receive higher rotation confidence, while nearby regions receive higher translation confidence. This behavior is consistent with the geometric motivation in Section~\ref{sec:method_conf}, indicating that pose supervision can learn different spatial preferences for rotational and translational constraints.

\begin{figure}[h]
    \centering
    \includegraphics[width=\linewidth]{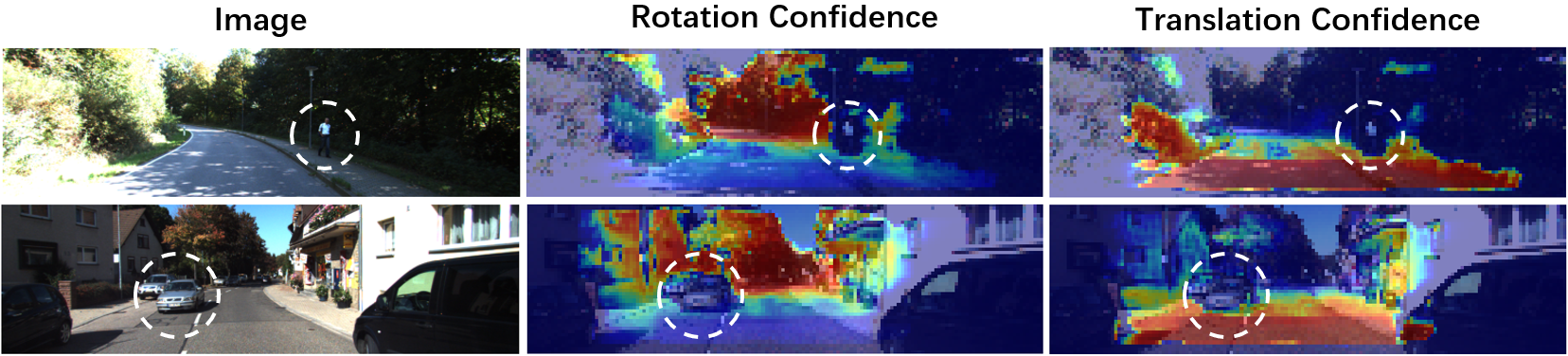}
    \caption{\textbf{Confidence prediction in dynamic regions.}
    SFVO assigns low confidence to dynamic objects, including pedestrians (top) and moving vehicles (bottom), reducing their influence on pose estimation.}
    \label{fig:conf_dyn}
\end{figure}

The confidence predictor also learns to suppresses correspondences inconsistent with static camera motion from pose supervision. As shown in Fig.~\ref{fig:conf_dyn}, dynamic objects such as pedestrians and moving vehicles receive noticeably lower confidence, reducing their influence on pose estimation.

To quantify the benefit of decoupling, we retrain SFVO with a shared confidence, and pose is estimated using weighted least-squares PnP. As reported in Table~\ref{tab:abl_decouple}, decoupled confidence consistently improves all pose metrics, with substantial reductions in both ATE and RPE.

Confidence decoupling also improves training efficiency. The decoupled model reaches its best performance within 5 epochs, whereas the coupled baseline requires more than 40 epochs to reach the reported accuracy. This suggests that separating the two confidence channels reduces conflicting weighting requirements between rotation and translation.

\begin{table}[h]
    \centering
    \caption{Ablation of Coupled and Decoupled Confidence on KITTI Sequences 09 and 10}
    \label{tab:abl_decouple}
    \renewcommand{\arraystretch}{1.2}
    \begin{tabular}{c c c c c}
        \hline
        Confidence & Sequence & ATE (m) & RPE$_t$ (m) & RPE$_r$ (deg) \\
        \hline
        Coupled & \multirow{2}{*}{09} & 15.006 & 0.053 & 0.111 \\
        Decoupled & & \textbf{9.830} & \textbf{0.016} & \textbf{0.039} \\
        \hline
        Coupled & \multirow{2}{*}{10} & 6.284 & 0.046 & 0.105 \\
        Decoupled & & \textbf{2.835} & \textbf{0.017} & \textbf{0.046} \\
        \hline
    \end{tabular}
\end{table}

\begin{figure}[h]
    \centering
    \includegraphics[width=\linewidth]{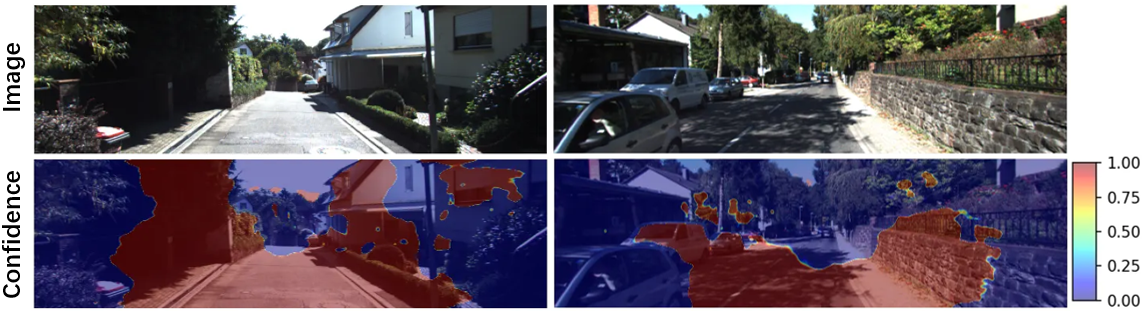}
    \caption{\textbf{Visualization of coupled confidence maps.}
    Sharing a single confidence map between rotation and translation produces highly polarized predictions and tends to suppress distant correspondences, despite their usefulness for rotational estimation.}
    \label{fig:conf_uni}
\end{figure}

The qualitative difference is further shown in Fig.~\ref{fig:conf_uni}. With coupled confidence, the predicted weights become highly polarized to 0 or 1, and frequently suppress distant regions. Since distant correspondences can be unreliable for translation while remaining informative for rotation, a shared confidence value cannot represent these two roles independently, leading to unnecessary rejection of useful rotational constraints.

\subsubsection{PnP Solver Convergence and Computation Cost}

\begin{figure}[h]
    \centering
    \includegraphics[width=\linewidth]{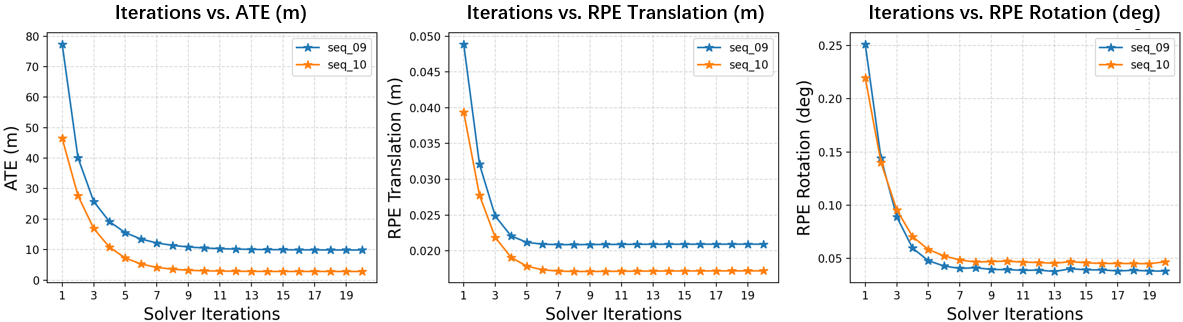}
    \caption{\textbf{Effect of pose solver iterations on KITTI 09.}
    With identity initialization, the pose estimate progressively converges and becomes stable after approximately 12 iterations.}
    \label{fig:abl_iterations}
\end{figure}

The presented PnP pose solver alternates between rotation and translation updates due to their geometric coupling. To evaluate convergence efficiency, we initialize the solver with $\mathbf{R}=\mathbf{I}$ and $\mathbf{t}=\mathbf{0}$. As shown in Fig.~\ref{fig:abl_iterations}, the pose estimate becomes stable after approximately 12 iterations.

The solver only involves lightweight batched linear algebra operations. On an NVIDIA RTX 4060 Ti, each iteration requires approximately $1.1\,\mathrm{ms}$. We use 20 iterations during training to ensure stable convergence from identity initialization. At inference time, using the previous frame-to-frame motion estimation result as a warm start reduces the convergence iterations to 7.

For stereo images of resolution $480\times640$, the peak GPU memory occupation is $930\,\mathrm{MB}$, the average end-to-end inference time is approximately $45\,\mathrm{ms}$ on an NVIDIA RTX 4090 and $130\,\mathrm{ms}$ on an RTX 4060 Ti, demonstrating practical computational efficiency across different GPU platforms.

\section{Conclusion}

In this paper, we present SFVO, a correspondence-driven deep stereo visual odometry framework that builds upon pretrained stereo matching and optical flow models. By directly converting disparity and temporal correspondences into dense 3D--2D geometric constraints, SFVO avoids learning low-level correspondence estimation from scratch and enables a simple and efficient frame-to-frame pose estimation pipeline. We introduce rotation--translation decoupled confidence weighting and a differentiable bidirectional PnP solver to improve the robustness of geometric pose estimation. Experiments on indoor and outdoor benchmarks demonstrate competitive accuracy, efficient inference, and strong cross-dataset generalization.

\bibliographystyle{IEEEtran}
\bibliography{references}

\end{document}